\documentclass[journal]{IEEEtai}

\usepackage[colorlinks,urlcolor=blue,linkcolor=blue,citecolor=blue]{hyperref}

\usepackage{color,array}

\usepackage{graphicx}

\usepackage{amsmath}
\usepackage{amssymb}
\usepackage{multirow}
\usepackage{makecell}

\usepackage{algorithm}
\usepackage{algorithmic}
\usepackage{cite}
\begin{document}

\title{I-Diff: Structural Regularization for High-Fidelity Diffusion Models} 
\author{Shakthi Perera, Dilum Fernando, H.L.P. Malshan, Herath Mudiyanselage Pathum Shehara Madushan, Roshan Godaliyadda, \IEEEmembership{Senior Member, IEEE}, Mervyn Parakrama Bandara Ekanayake, \IEEEmembership{Senior Member, IEEE}, Dhananjaya Jayasundara, and Roshan Ragel, \IEEEmembership{Senior Member, IEEE}}
% \thanks{This work was supported in part by NVIDIA Academic Grant Program.}
% \thanks{This paragraph of the first footnote will contain the date on which you submitted your paper for review. It will also contain support information, including sponsor and financial support acknowledgment. For example, ``This work was supported in part by the U.S. Department of Commerce under Grant BS123456.'' }
% \thanks{The next few paragraphs should contain the authors' current affiliations, including current address and e-mail. For example, F. A. Author is with the National Institute of Standards and Technology, Boulder, CO 80305 USA (e-mail: author@boulder.nist.gov).}
% \thanks{S. B. Author, Jr., was with Rice University, Houston, TX 77005 USA. He is now with the Department of Physics, Colorado State University, Fort Collins, CO 80523 USA (e-mail: author@lamar.colostate.edu).}
% \thanks{T. C. Author is with the Electrical Engineering Department, University of Colorado, Boulder, CO 80309 USA, on leave from the National Research Institute for Metals, Tsukuba, Japan (e-mail: author@nrim.go.jp).}
% \thanks{This paragraph will include the Associate Editor who handled your paper.}

% \markboth{Journal of IEEE Transactions on Artificial Intelligence, Vol. 00, No. 0, Month 2020}
% {First A. Author \MakeLowercase{\textit{et al.}}: Bare Demo of IEEEtai.cls for IEEE Journals of IEEE Transactions on Artificial Intelligence}
\maketitle

\begin{abstract}
Denoising Diffusion Probabilistic Models (DDPMs) have significantly advanced generative AI, achieving impressive results in high-quality image and data generation. However, enhancing fidelity without compromising semantic content remains a key challenge in the field. Recent diffusion research in multiple disciplines has introduced objectives and architectural refinements that tighten the match between generated and real data distributions, yielding higher fidelity than earlier generative frameworks. Multi-stage architectures, physics-guided modeling, semantic conditioning, and rarity-aware generation are some of the explored works to achieve this task. However, the integration of structural information of the data distribution into DDPM has largely been unexplored. The conventional DDPM framework relies solely on the $L^2$ norm between the additive and predicted noise to generate new data distributions. We introduce I-Diff, an improved version of DDPM that incorporates a carefully designed regularizer, effectively enabling the model to encode structural information, thereby preserving the inherent fidelity of the data distribution. The proposed approach is validated through extensive experiments on DDPM, Improved DDPM and Latent Diffusion Model across multiple datasets. Empirical results demonstrate significant improvements in fidelity (Density and Precision increase 10\% and 37\% in CIFAR-100 dataset respectively) across other tested datasets. These results highlight the effectiveness of our method in enhancing the fidelity of the generated data. Notably, improvements across different models highlight the model-agnostic nature of our proposed method in the wider field of generative AI.
\end{abstract}

\begin{IEEEImpStatement}
High-fidelity generation is critical in multidisciplinary domains such as medical imaging, remote sensing, and speech synthesis, where even subtle deviations in synthesized data can lead to significant real-world consequences. Such errors can accumulate through recursive reuse and propagate systematic bias across subsequent learning pipelines. These applications demand generative models that respect the underlying semantics of the true data distribution precisely rather than producing superficially realistic outputs. This work introduces a simple, computationally efficient modification that can be integrated into any DDPM variant to enhance fidelity across such domains without imposing architectural or domain-specific constraints. The approach strengthens alignment with real data while remaining broadly applicable and inexpensive to deploy. In addition, the study underscores the value of Precision, Recall, Density, and Coverage metrics alongside FID and Inception Score, providing a more interpretable assessment of fidelity and diversity in generative settings.
\end{IEEEImpStatement}


\section{Introduction}

\IEEEPARstart{F}{idelity} in data generation captures the degree to which generated data accurately preserve the semantic structure and statistical characteristics of real-world data. High-fidelity generation has become a critical objective in the modern generative paradigm, given the increasing reliance on synthetic data for downstream tasks where even subtle deviations can lead to misinterpretation or degraded performance. For instance, the growing use of synthetic data in AI model training \cite{dataugmentationframe, augdiff} implies that any imprecision in the generated samples may be embedded in the learned representations, thereby amplifying downstream errors through a cascading effect. This risk is particularly acute in safety-critical domains such as healthcare, where imprecise synthetic medical data can lead to false positives and misleading clinical inferences. Furthermore, the importance of high-fidelity data generation is particularly evident in recent research across domains such as medical imaging \cite{martel_spatial-intensity_2020}, speech synthesis \cite{kong_hifi-gan_2020}, and remote sensing \cite{li_high-fidelity_2024}, where realism and statistical consistency are essential. Generative modeling frameworks, including Variational Autoencoders (VAEs) \cite{kingma_auto-encoding_2013}, Generative Adversarial Networks (GANs) \cite{goodfellow_generative_2020}, flow-based models \cite{rezende_variational_2015}, and energy-based methods \cite{du_implicit_2019} have contributed significantly toward this goal. Nonetheless, diffusion models \cite{ho_denoising_2020, sohl-dickstein_deep_2015} have proven particularly effective for high-fidelity data generation, often surpassing alternatives in sample quality, making them a cornerstone for a wide variety of applications.

Diffusion models have been accomplishing great feats in the realm of generative AI, specifically in terms of unconditional and conditional image generation. Starting with the revolutionary paper Denoising Diffusion Probabilistic Model (DDPM) by Ho \textit{et al.} (2020) \cite{ho_denoising_2020} and the improvements by Nichol \& Dhariwal (2021) \cite{dhariwal_diffusion_2021} as well as the Latent Diffusion Model (LDM) by Rombach \textit{et al.} (2022) \cite{rombach_high-resolution_2021}, these models have had the biggest impact in this context. The fidelity and diversity of the images generated by these models are surprisingly amazing. Furthermore, the growing body of research directly targeting fidelity using diffusion models \cite{ho_cascaded_2021, sehwag_generating_2022, shu_physics-informed_2023, dorjsembe_conditional_2024} provides clear evidence that improving fidelity remains a crucial objective in diffusion based generative modeling as well.

In the context of improving DDPMs \cite{ho_denoising_2020}, there have been several works contributed to advance DDPMs beyond their original formulation. Improved diffusion models \cite{nichol_improved_2021} introduced techniques such as cosine-based variance schedules to enhance training stability and sample quality. Variational diffusion models further extended this direction by proposing a unifying framework with refined objectives. Denoising Diffusion Implicit Models (DDIM) \cite{song_denoising_2020} provided a non-Markovian formulation that allows deterministic sampling, while Pro-DDPM \cite{salimans_progressive_2022} accelerated generation through progressive distillation. Complementing these methods, the Min-SNR weighting \cite{hang_efficient_2023} strategy improved training efficiency by adapting the importance assigned to different noise levels.

\begin{figure*}[!t]
\centering
\includegraphics[width=1.0\linewidth]{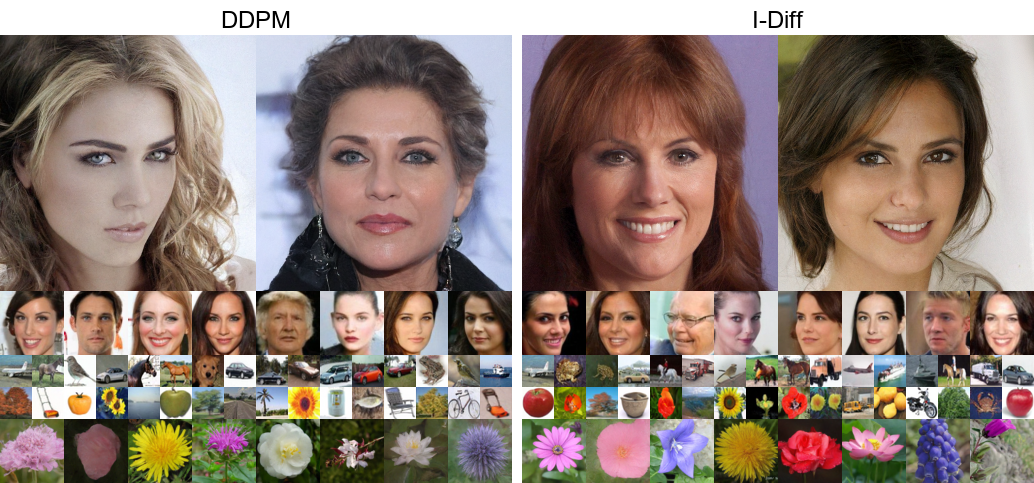}
\vspace{-3mm}
\caption{Comparison of generated images from DDPM (left) and I-Diff (right) across CelebA-HQ, CelebA, CIFAR-10, CIFAR-100, and Oxford Flowers (top to bottom).}
\label{fig:generated_images}
\end{figure*}

More recent research has focused on modifying the noise design in the forward diffusion process. Blue-noise diffusion \cite{huang_blue_2024} replaced standard Gaussian noise with structured blue-noise patterns to produce perceptually sharper generations. Edge-preserving diffusion models \cite{vandersanden_edge-preserving_2024} introduced noise schemes that emphasize boundaries and fine details. Extending these ideas to non-Euclidean domains, directional diffusion models \cite{yang_directional_2023} applied noise injection strategies tailored to graph structures, enabling more expressive representation learning and recommendation \cite{yi_directional_2024}.

As the core model utilized in all the aforementioned works, in DDPMs, the forward process gradually destroys the structure of data by adding noise until it becomes an isotropic Gaussian. The reverse process then learns to reconstruct the original structure step by step, effectively modeling how structured data emerges from randomness. Even though the aforementioned approaches highlight that ongoing research continues to seek further improvements at the fundamental level of diffusion design, incorporating structural information to DDPM remains largely absent. In the DDPM implementation, the learning process considers the expected squared norm ($L^2$) difference between the additive Gaussian noise and the predicted noise as its objective function. Therefore, for the generative process, to enhance the aforementioned creation of structure, the objective function can be modified to include any structural measure. Our assertion is that by capturing the statistical properties of the noise in more detail, the model will be able to produce higher-fidelity samples as it would have much more information regarding the distributional structure of the samples.

To address this, we compare several structural measures and choose the best performing and computationally efficient measure as the optimal measure. We experiment on four well-defined 2D synthetic datasets, such as Swiss Roll, Scattered Moon, Moon with Two Circles, and Central Banana, drawing fundamental conclusions about the DDPM algorithm. Finally, we validate our approach to unconditional image generation using the Oxford Flower \cite{nilsback_automated_2008}, Oxford-IIITPet \cite{parkhi_cats_2012}, CIFAR-10 \cite{krizhevsky_cifar-10_nodate}, CIFAR-100 \cite{krizhevsky_cifar-10_nodate}, CelebA \cite{liu2015faceattributes}, and CelebA-HQ \cite{karras2018progressive}datasets for baseline DDPM, improved DDPM, and LDM.

The contributions of this work are as follows:
\begin{itemize}
    \item We introduce I-Diff: a modified approach that introduces a structural constraint on the predicted noise objective function to steer the generative process in a structurally coherent manner. This results in improved fidelity of the generated data distribution. To the best of our knowledge, we are the first to propose such a modified loss based on the structural properties of the noise. 

    \item We evaluate and validate our approach on four 2D synthetic datasets as well as on the task of unconditional image generation on Oxford Flower \cite{nilsback_automated_2008}, Oxford-IIIT Pet \cite{parkhi_cats_2012}, CIFAR-10 \cite{krizhevsky_cifar-10_nodate}, CIFAR-100 \cite{krizhevsky_cifar-10_nodate}, CelebA \cite{liu2015faceattributes}, and CelebAHQ \cite{karras2018progressive}  datasets. Considering the key evaluation metrics, such as Precision and Recall \cite{kynkaanniemi_improved_2019}, Density and Coverage \cite{naeem_reliable_2020}, Frechet Inception Distance (FID) \cite{heusel_gans_2017} and Inception Score (IS) \cite{salimans_improved_2016}, the modified objective is able to surpass the original DDPM with a significant gap in terms of the fidelity metrics, Density and Precision.

    \item We conduct an in-depth analysis of the Density and Coverage metrics to evaluate the generative capabilities of I-Diff compared to DDPM. This analysis facilitates a detailed comparison between the generated and true data distributions, visually illustrating I-Diff’s superior alignment with the true distribution. Furthermore, it provides valuable insights on the importance of analyzing the performance of generative AI algorithms on these fidelity measures for numerous applications.  
\end{itemize}

\section{Background}
\subsection{Definitions}
In the DDPM, we simply add a Gaussian noise, which varies according to a specific variance schedule $\beta_t \in (0, 1)$. The noise at each time step corrupts the data, such that by the time the time step reaches its final value $T$, the data will be mapped to an almost isotropic Gaussian distribution. However, the learning occurs when we try to learn the reverse process by which we try to denoise along the same trajectory starting from the almost isotropic Gaussian distribution. The first process, in which we add noise, is called the forward process and the latter, in which we denoise, is called the reverse process. The forward process is often characterized by $q$ and the reverse process by $p$. Both of which are modeled as Gaussian distributions. 

The forward process is defined as follows,
\begin{equation}
  q(x_1,x_2, \dots x_T|x_0) \triangleq \prod_{t=1}^{T} q(x_t | x_{t-1}) \\
  \label{eq:1}
\end{equation}
\begin{equation}
  q(x_t | x_{t-1}) \triangleq \mathcal{N}(x_t; \sqrt{1 - \beta_t}x_{t-1}, \beta_t\mathbf{I}) 
  \label{eq:2}
\end{equation}

Moreoever, by introducing $ \alpha_t = 1 - \beta_t$ as well as $\bar{\alpha_t} = \prod_{i=1}^t \alpha_i$ the forward process can be further simplified into the following expression via the re-parametrization trick \cite{kingma_auto-encoding_2013}. 

\begin{equation}
    q(x_t | x_{0}) = \mathcal{N}(x_t; \sqrt{\bar{\alpha_t}}x_{0}, (1- \bar{\alpha_t})\mathbf{I})
    \label{eq:4}
\end{equation}
\begin{equation}
    x_t = \sqrt{\bar{\alpha_t}}x_0 + \sqrt{1-\bar{\alpha_t}}\epsilon 
    \label{eq:5}
\end{equation}
where, $\epsilon \sim \mathcal{N}(0, \mathbf{I})$. 

The reverse process, given by $p \sim \mathcal{N}(x_{t-1} | x_t)$, can be obtained in terms of the forward process distribution $q$ and Baye's Theorem. However, the reverse process only becomes tractable when the posterior distribution $ q(x_{t-1} | x_t)$, is conditioned on the input data $x_0$. Thus, during training, the model tries to learn the tractable $  q(x_{t-1} | x_t, x_0) $ distribution. This distribution, which is also a Gaussian distribution, is defined by the following equation and parameters.
\begin{equation}
    q(x_{t-1} | x_t, x_0) = \mathcal{N}(x_{t-1};\Tilde{\mu}(x_t, x_0), \Tilde{\beta}_t\mathbf{I})
    \label{eq:6}
\end{equation}
\begin{equation}
    \Tilde{\beta}_t = \frac{1 - \bar{\alpha}_{t-1}}{1-\bar{\alpha}_t}\beta_t
    \label{eq:7}
\end{equation}
\begin{equation}
    \Tilde{\mu}_t(x_t, x_0) = \frac{\sqrt{\bar{\alpha}_{t-1}}\beta_t}{1-\bar{\alpha}_t}x_0 +  \frac{\sqrt{{\alpha}_t}(1 - \bar{\alpha}_{t-1})}{1-\bar{\alpha}_t}x_t
    \label{eq:8}
\end{equation}

\subsection{Training Process}
To train, however, one could make the model predict the mean of the reverse process distribution at each time step. But Ho \textit{et al.} (2020) \cite{ho_denoising_2020} mentions that predicting the additive noise, $\epsilon$, leads to better results. The additive noise and the mean of the reverse process distribution at each time step are elegantly linked by (\ref{eq:5}) and (\ref{eq:8}). This results in the following re-parametrization of $\Tilde{\mu}(x_t, t)$,
\begin{equation}
    \Tilde{\mu}(x_t, t) = \frac{1}{\sqrt{\alpha_t}}\left(x_t - \frac{1 - \alpha_t}{\sqrt{1-\bar{\alpha}_t}}\epsilon\right)
    \label{eq:9}
\end{equation}
Therefore, predicting the additive noise $\epsilon$, is adequate for the task of predicting the mean of the backward process distribution. Moreover, since the forward process' variance schedule is fixed, the reverse process variance, $\Tilde{\beta}_t$, is also assumed to be fixed according to $\Tilde{\beta}_t$. 
Thus, Ho \textit{et al.} (2020) \cite{ho_denoising_2020} proposes to optimize the following simple objective function during the training process. 
\begin{equation}
    L_{\text{simple}} = \mathbb{E}_{t,x_0,\epsilon}[||\epsilon - \epsilon_{\theta}(x_t, t)||^2]
    \label{eq:10}
\end{equation}
where $\epsilon_{\theta}(x_t, t)$ is the predicted noise.
\subsection{Why a Structural Regularizer}
\label{sec:why_regularizer}

\begin{figure*}[!t]
\centering
\includegraphics[width=1.0\linewidth]{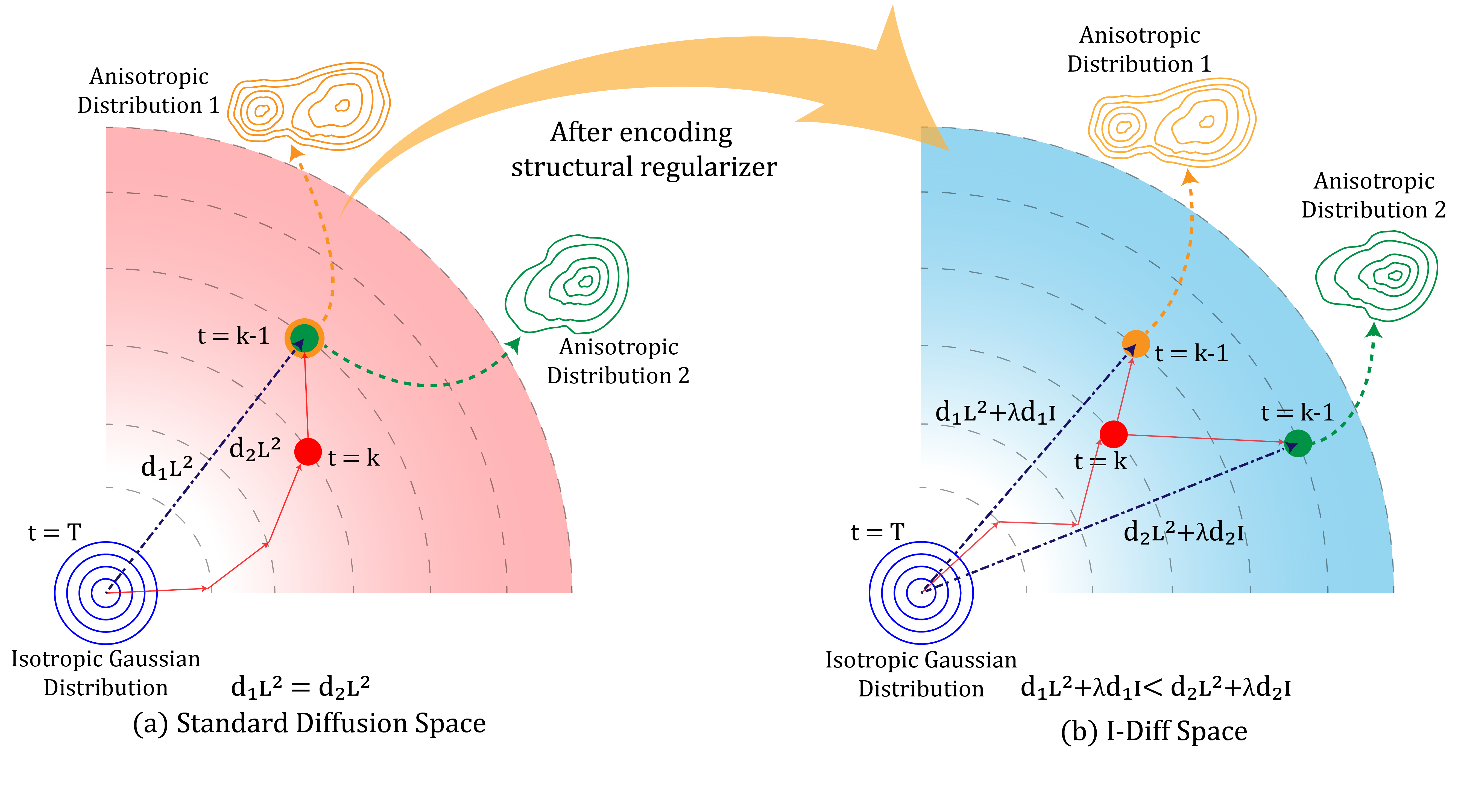}
\vspace{-10mm}
\caption{\textbf{How the structural regularizer guides the model toward a more structurally coherent data distribution compared to the reverse process of the standard DDPM}: The left side of the figure illustrates that during the reverse diffusion process, when the model encounters equidistant distributions at a particular time step, it is not explicitly guided to select the most structurally coherent distribution, leading to a mismatch with the true underlying data distribution. On the right side, when the structural constraint (I-Diff) is applied, previously equidistant distributions are now separated by distinct distances. This explicit structural enforcement enables the reverse diffusion process to steer towards a distribution that more accurately aligns with the true data distribution.}
\label{fig:why_regularizer}
\end{figure*}

Let us reformulate the diffusion framework on a metric space to reimagine its underlying mechanisms visually. Consider a metric space where each point represents a distribution, and the distance between points is measured using the $L^2$ norm. Furthermore, we assume that the origin of this space corresponds to an isotropic Gaussian distribution ($g(\cdot)$), which reflects the structural vanishing point of the forward diffusion process.
Mathematically, for a given predicted noise distribution, the distance from the origin is
\vspace{-2mm}
\begin{equation}
d_{L^2}(g_{\theta, t}(\cdot), g(\cdot), t) = \left( \int (g_{\theta, t}(y) - g(y))^2 \, dy \right)^{1/2}
\label{eq:a}
\end{equation}
Here, $g_{\theta, t}(\cdot)$ is the probability distribution function of the predicted noise ($\epsilon_\theta$) for a given data point $x_t$ at time sample $t$. It is obtained as the solution to the minimization problem defined in (\ref{eq:10}). This distance quantifies how far a given distribution is from an isotropic Gaussian distribution. The reverse diffusion process can thus be seen as a path in this space, starting at the origin and moving toward the desired distribution, with the $L^2$ norm providing a metric structure. For this to be a metric space, the distance must satisfy certain conditions mentioned and derived in the Supplementary Materials Section A. The original DDPM algorithm, at the $k^{th}$ step of the reverse process, predicts a noise distribution by learning how to map adjacent points in a metric space that represents the difference between $x_k$ and $x_{k-1}$ (as in (\ref{eq:10})). However, this formulation is limited to learning only the $L^2$ norm disparity. Consequently, if two distributions share the same $L^2$ norm but differ in a structural manner within this metric space, the algorithm remains agnostic to those differences (See Fig. \ref{fig:why_regularizer} (a)). Building upon this, if the transition from $q(x_k)$ to $q(x_{k-1})$ is defined in terms of an $L^2$ norm, multiple structurally different distributions with the same $L^2$ norm may exist at the same point in this space. As a result, an $L^2$-norm-based metric space alone may fail to distinguish structural differences between the generated data distributions. Under this formulation, the reverse process iteratively refines $q(x_k)$ to $q(x_{k-1})$ until the desired final distribution, $q(x_0)$ is generated, which might be handicapped by the aforementioned limitations.

To address this, incorporating a statistical measure derived from properties satisfied at the global minimum into the metric space may allow the model to account for both $L^2$ norm disparities and structural variations. This enables the predicted noise to encode structural differences more effectively. Under the assumption of a complex loss landscape with many local minima, the additional regularizer helps guide the optimizer toward more effective directions, reducing the likelihood of suboptimal solutions \cite{Goodfellow-et-al-2016}. Mathematically, the distance in the metric space after encoding the structural regularizer (I-Diff Space) is,
\begin{equation}
d_{\text{new}}(g_{\theta, t}(\cdot), g(\cdot), t) = g_{L^2}(g_{\theta, t}(\cdot), g(\cdot), t) + \lambda \cdot d_{I}(g_{\theta, t}(\cdot), g(\cdot), t)
\label{eq:b}
\end{equation}
Here, $d_{L^2}(g_{\theta, t}(\cdot), g(\cdot), t)$ denotes the default $L^2$ norm distance, and $d_{I}(g_{\theta, t}(\cdot), g(\cdot), t)$ denotes the statistical measure that captures the disparity in the predicted noise for a given $x_t$ at time sample $t$ compared to the isotropic Gaussian distribution $g(\epsilon)$, and $\lambda$ is the regularization parameter. It can be shown that I-Diff space becomes a metric space as proved in Supplementary Materials Section A.
As a result, after the noise adjustment, the generated data distributions reflect not only $L^2$ norm-based deviations but also structural disparities. In this revised metric space, two distributions with identical $L^2$ norms but distinct structural properties can be mapped to separate points, as illustrated in Fig. \ref{fig:why_regularizer}, thereby facilitating enhanced structural encoding on the generated data distribution. In this structurally enriched metric space, the final distribution, $q(x_0)$ is recursively formed by transitioning from $q(x_k)$ to $q(x_{k-1})$ while imposing statistical properties of the distribution. This process inherently provides a greater capability to capture the complex geometries of the data distribution, which is reflected in the results, even with a computationally inexpensive amendment to the objective function.

\section{Methodology}
In the default DDPM, the variance schedule drives the distribution toward an isotropic Gaussian distribution by restricting the degrees of freedom for the movement of information of the distribution, without using backpropagation to adaptively learn the degree of statistical property achieved, making it, a non-learnable process. With the identified need for a structural regularizer, mean, skewness \cite{mardia_measures_1970}, kurtosis \cite{mardia_measures_1970}, Kullback-Liebler (KL) Divergence, Maximum Mean Discrepancy (MMD) \cite{JMLR:v13:gretton12a}, and isotropy were utilized as defined below. 

Let \(\epsilon_{\theta} \in \mathbb{R}^d\) be a predicted noise vector with distribution \(p\) and  $\Sigma_\epsilon$ be the covariance of $\epsilon_{\theta}$, 

\textbf{Mean:}  
\[
\mu_\epsilon = \mathbb E[\epsilon_{\theta}]
\qquad
\]

\textbf{Skewness:}
\[
\gamma_{1,d}
= \mathbb E\!\left[
\left(
(\epsilon_{\theta} - \mu_\epsilon)^\top
\Sigma_\epsilon^{-1}
(\epsilon_{\theta} - \mu_\epsilon)
\right)^{3/2}
\right].
\]

\textbf{Kurtosis :}
\[
\gamma_{2,d}
= \mathbb E\!\left[
\left(
(\epsilon_{\theta} - \mu_\epsilon)^\top
\Sigma_\epsilon^{-1}
(\epsilon_{\theta} - \mu_\epsilon)
\right)^{2}
\right].
\]

\textbf{Kullback–Leibler divergence:}  
For distributions \(p\) and \(q\):  
\[
D_{\mathrm{KL}}(p\Vert q)
= \int p(x)\,\log\frac{p(x)}{q(x)}\,\mathrm{d}x.
\]

\textbf{Maximum Mean Discrepancy (MMD):} 
For kernel \(k\) and feature map \(\phi\colon \mathcal X \to \mathcal H\):  
\[
\mathrm{MMD}^2(p,q)
= \big\|\mathbb E_p[\phi(\epsilon_{\theta})] - \mathbb E_q[\phi(\epsilon'_{\theta})]\big\|_{\mathcal H}^2.
\]

\noindent Here, $\mathcal X$ denotes the input space, and $\mathcal H$ denotes the reproducing-kernel Hilbert space (RKHS) associated with the kernel $k$.

\textbf{ISO-1:Iso Trace Mean}  
\[
{\mathrm{iso}\mbox{-}1} = \mathbb E \bigl(\epsilon_{\theta}^{\!\top}\epsilon_{\theta}\bigr)
\]

\textbf{ISO-2: Frobenius Norm Deviation}  
\[
{\mathrm{iso}\mbox{-}2}
= \big\|\Sigma_\epsilon - I_d\big\|_F^2.
\]

\textbf{ISO-3: Diagonal–Off-Diagonal Split}  
\[
{\mathrm{iso}\mbox{-}3}
= \frac1{d(d-1)}\sum_{i\neq j}\Sigma_{\epsilon_{ij}}^2 \;+\; \Big(\frac1d\,\mathrm{Tr}(\Sigma_\epsilon) - 1\Big)^2.
\]

\textbf{ISO-4: Log-Eigenvalue Penalty}  

Let \(\{\lambda_k\}_{k=1}^d\) be the eigenvalues of \(\Sigma_\epsilon\).  
\[
{\mathrm{iso}\mbox{-}4}
= \frac1d \sum_{k=1}^d (\log \lambda_k)^2.
\]

\textbf{ISO-5: Bures Distance Penalty}  
\[
{\mathrm{iso}\mbox{-}5}
= \frac1d \sum_{k=1}^d \big(\sqrt{\lambda_k} - 1\big)^2.
\]

Among these measures, iso trace mean consistently performs well compared to the others when incorporated into the loss function as will be shown in the Results and Discussion section. In particular, it leads to marked improvements in fidelity-related metrics such as Precision and Density. The new modified objective function we propose to optimize is,

\begin{equation}
    L_{\text{modified}} = \mathbb{E}(||\epsilon - \epsilon_{\theta}||^2) + \lambda (\mathbb{E}(\epsilon_{\theta}^T\epsilon_{\theta}) - n)^2 \label{eq:31}
\end{equation}
where $\lambda$ is the regularization parameter and $n$ denotes the isotropy of an isotropic random vector in $\mathbb{R}^n$ is the expected squared norm of that vector, which is equal to its dimension \cite{Goodfellow-et-al-2016}.

However, this modified objective needs to be further simplified so as to make this new error independent of the size of the dimension of the random vector. Thus, we introduce the following normalization during implementation to ensure that the loss remains dimension-invariant.

\begin{equation}
    L_{\text{modified}} = \mathbb{E}(||\epsilon - \epsilon_{\theta}||^2) + \lambda \left(\mathbb{E}\left(\frac{\epsilon_{\theta}^T\epsilon_{\theta}}{n}\right) - 1\right)^2 \label{eq:32}
\end{equation}

\section{Experiments}

\subsection{Implementation Details}  
To compare the performance between DDPM and I-Diff, the modified loss function was utilized in all four 2D synthetic datasets. Precision, Recall, Density along with Coverage were used to evaluate and compare the performance of the two models on 2D synthetic datasets. In addition to those four evaluation metrics, FID and IS were used to evaluate the quality of the generated samples for the image datasets Oxford Flower, Oxford-IIIT Pet, CIFAR-10, CIFAR-100, CelebA, and CelebAHQ. Moreover,  Improved DDPM and Latent Diffusion Models (LDMs) were trained on CIFAR-10, CIFAR-100, and CelebA. All training configurations, hyperparameter selections, and implementation details for the diffusion models used in this work are provided in the Supplementary Materials Section B.

\begin{figure}[!b]
\centering
\includegraphics[width=1.0\linewidth]
{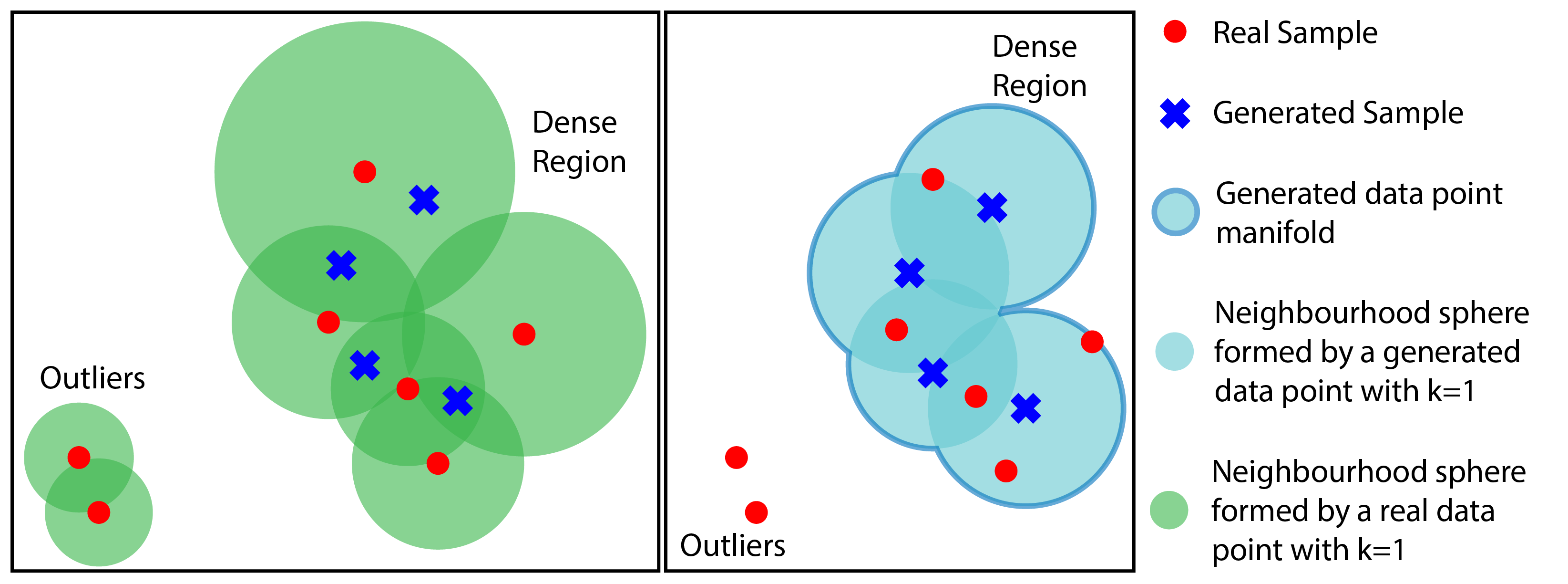}
\caption{An example scenario for illustrating a situation where high Density and low Coverage is recorded. Generating samples in the neighborhoods of the highly dense regions over the  outliers in the true manifold has resulted in a high Density and low Coverage.}
\label{fig:prdc_interpretation}
\end{figure}

\subsection{Evaluation Metrics}
\textbf{Fréchet Inception Distance (FID)} measures the similarity between real and generated data distributions in the feature space of a pretrained Inception network. 
After extracting features, each distribution is modeled as a multivariate Gaussian with means $\mu_r, \mu_g$ and covariances $\Sigma_r, \Sigma_g$. 
The FID score is computed as
\[
\mathrm{FID} 
= \|\mu_r - \mu_g\|_2^2
+ \mathrm{Tr}\!\left(
\Sigma_r + \Sigma_g 
- 2\sqrt{\Sigma_r\Sigma_g}
\right).
\]
Lower FID indicates that the generated samples more closely match the real distribution in terms of their feature statistics and global structure.

% \bigskip

\textbf{Inception Score (IS)} evaluates both the quality and diversity of generated samples using the predictions of a pretrained classifier. 
It is defined as
\[
\mathrm{IS} = 
\exp\!\left(
\mathbb{E}_{x \sim p_g}\!
\left[
\mathrm{KL}\big( p(y|x) \,\|\, p(y) \big)
\right]
\right),
\]
where $p(y|x)$ is the conditional label distribution for each generated sample, and $p(y)$ is the marginal label distribution over all generated samples.
Higher IS values indicate (i) confident, low-entropy predictions for individual samples, and (ii) a diverse set of generated samples spanning many predicted classes.

\begin{figure*}[!t]
\centering
\includegraphics[width=1.0\linewidth]
{FiguresPaper/perer4.png}
\caption{Percentage difference in Precision and Density relative to baseline DDPM for Central Banana, Moon Circles, Scattered Moon, and Swiss Roll datasets (left to right), evaluated under different statistical measures as regularizers}
\label{fig:objectiveFunction}
\end{figure*}

\textbf{Precision, Recall, Density, and Coverage}\\
Let us define a couple of terms that would be useful to understand the evaluation metrics. A neighborhood sphere is the local area around a data point that includes nearby points, defined by a radius that is the distance to the k-th nearest neighbor (see Fig. \ref{fig:prdc_interpretation}). True manifold is created with the collection of neighborhood spheres formed by real data points. Similarly, a generated sample manifold is created with the collection of neighborhood spheres formed by generated data points.

\textbf{Precision} denotes the fraction of generated data that lies in the true manifold by counting whether each generated data point falls within a neighborhood sphere of real samples. This measure reflects how closely the generated points align with the true manifold \cite{kynkaanniemi_improved_2019, sajjadi_assessing_2018}.
In the left of Fig. \ref{fig:prdc_interpretation}, all four generated points lie within the true manifold; therefore, Precision is 4/4.

\textbf{Recall} denotes the fraction of true data that lies in the generated sample manifold. This measure indicates how well the true points align with the generated sample manifold \cite{kynkaanniemi_improved_2019, sajjadi_assessing_2018}. However, the lack of generated samples near sparse outliers in the true data leads to low Recall, as the sample manifold fails to capture these regions. In the right of Fig. \ref{fig:prdc_interpretation}, five out of seven real samples are covered; therefore, Recall is 5/7.

\textbf{Density} counts the number of neighborhood spheres of real samples that encompass each generated data point. This allows Density to reward generated samples located in areas densely populated by real samples, reducing sensitivity to outliers. This enables us to consider the local context of a distribution by measuring how close a sample is to densely packed points in the true distribution \cite{naeem_reliable_2020}. In the left of Fig. \ref{fig:prdc_interpretation}, summing the overlaps, which are 3, 2, 2, and 1; therefore, Density equals 8/7.

\textbf{Coverage} measures the fraction of real samples whose neighborhoods contain at least one generated sample.  Moreover, Coverage measures the diversity of a distribution by assessing whether all aspects of the distribution are represented. However, the presence of sparse outliers in the true manifold and the absence of the generated samples near those outliers may result in low Coverage \cite{naeem_reliable_2020}. In the left of Fig. \ref{fig:prdc_interpretation}, five out of seven real samples have a generated sample nearby; therefore, Coverage is 5/7.

\section{Results and Discussion}
\subsection{Comparison of regularizers as structural measures}
Across all experimental settings, the results in Supplementary Materials Section C (Tables IV, V, VI, and VII) exhibit a clear fidelity–diversity trade-off introduced by adding structural regularizers to the original diffusion objective. Methods that enforce stronger structural constraints consistently increase fidelity (Precision, Density) while reducing diversity (Recall, Coverage). Among all evaluated regularizers, the iso trace mean objective is the only method that yields uniform improvements in both Precision and Density (see Fig. \ref{fig:objectiveFunction}) across all datasets while maintaining an acceptable decline in diversity.

Trace-based measure also outperforms mean measure and higher-order moment statistics such as kurtosis and skewness. Although higher-order diagnostics are more expressive in principle, they do not translate into better generative fidelity in practice. Computationally heavier isotropy-enforcing objectives such as eigenvalue-based measures, the Frobenius-norm penalty, and the Diagonal-Split covariance constraint explicitly push the predicted noise toward an identity covariance. However, the simple trace-mean regularizer still achieves higher Precision and shows more stable Density gains.

More sophisticated distance measures, including MMD with Riemannian kernel Hilbert spaces and KL-based penalties, show no consistent advantage. Despite their theoretical expressiveness, these objectives fail to surpass the trace-mean regularizer in fidelity-oriented metrics. The empirical evidence therefore supports selecting iso trace mean as the most effective and computationally efficient objective for all subsequent experiments. 

\begin{table*}
\centering
\renewcommand{\arraystretch}{1.3}
\caption{Evaluation of DDPM vs I-Diff for Oxford Flowers, Oxford-IIIT-Pet, CIFAR10, CIFAR100, CelebA, CelebAHQ datasets}
\begin{tabular}{c|c|cccccc}
\hline
\multirow{2}{*}{Dataset} & \multirow{2}{*}{Model} & \multicolumn{6}{c}{Evaluation Metrics} \\
\cline{3-8}
 &  & FID ($\downarrow$) & IS ($\uparrow$) & Precision ($\uparrow$) & Recall ($\uparrow$) & Density ($\uparrow$) & Coverage ($\uparrow$) \\
\hline \hline
\multirow{2}{*}{Oxford Flower} & DDPM & \textbf{30.4751} & 4.2619 & 0.9846 & \textbf{0.1117} & 23.8390 & 0.9999 \\
& $\text{I-Diff}_{\text{DDPM}}$
 & 32.0068 & \textbf{4.3039} & \textbf{0.9850} & 0.1087 & \textbf{28.4171} & 0.9999 \\
\hline
\multirow{2}{*}{Oxford-IIIT-Pet} & DDPM & 50.6720 & \textbf{6.6735} & 0.6599 & \textbf{0.3101} & 1.5205 & 0.9476 \\
& $\text{I-Diff}_{\text{DDPM}}$
 & \textbf{48.1217} & 6.4674 & \textbf{0.7652} & 0.2478 & \textbf{2.6284} & \textbf{0.9893} \\
\hline
\multirow{2}{*}{CIFAR-10} & DDPM & \textbf{3.4428} & 9.2835 & 0.6056 & \textbf{0.5320} & 1.0901 & 0.9704 \\
& $\text{I-Diff}_{\text{DDPM}}$
 & 3.6445 & \textbf{9.4049} & \textbf{0.6609} & 0.4849 & \textbf{1.4268} & \textbf{0.9851} \\
\hline
\multirow{2}{*}{CIFAR-100} & DDPM & \textbf{4.9657} & \textbf{10.4386} & 0.5901 & \textbf{0.5226} & 1.0620 & 0.9649 \\
& $\text{I-Diff}_{\text{DDPM}}$
 & 5.5034 & 10.2881 & \textbf{0.6482} & 0.4705 & \textbf{1.4578} & \textbf{0.9865} \\
\hline
\multirow{2}{*}{CelebA} & DDPM & \textbf{3.5738} & \textbf{2.3873} & 0.7309 & \textbf{0.6237} & 1.1329 & 0.9686 \\
& $\text{I-Diff}_{\text{DDPM}}$
 & 4.1475 & 2.3736 & \textbf{0.7589} & 0.6029 & \textbf{1.2736} & \textbf{0.9778} \\
\hline
\multirow{2}{*}{CelebAHQ} & DDPM & 14.7246 & \textbf{2.8276} & 0.7653 & 0.3778 & 1.9512 & 0.9659 \\
& $\text{I-Diff}_{\text{DDPM}}$
 & \textbf{14.1875} & 2.8097 & \textbf{0.7804} & \textbf{0.4064} & \textbf{2.1143} & \textbf{0.9824} \\
\end{tabular}
\label{table:1-DDPM_vs_IDiff}
\end{table*}

\begin{table}
\centering
\setlength{\tabcolsep}{2pt}
\renewcommand{\arraystretch}{1.3}
\caption{Evaluation of Improved DDPM vs $\text{I-Diff}_{\text{Imp.}}$ for CIFAR10, CIFAR100, CelebA datasets}

\begin{tabular}{c|cc|cc|cc}
\hline
\multirow{2}{*}{\makecell[c]{Evaluation\\Metrics}} & \multicolumn{2}{c|}{CIFAR-10} & \multicolumn{2}{c|}{CIFAR-100} & \multicolumn{2}{c}{CelebA} \\
\cline{2-7}
& \makecell[c]{Improved\\DDPM} & $\text{I-Diff}_{\text{Imp.}}$ & \makecell[c]{Improved\\DDPM} & $\text{I-Diff}_{\text{Imp.}}$ & \makecell[c]{Improved\\DDPM} & $\text{I-Diff}_{\text{Imp.}}$ \\
\hline
FID ($\downarrow$) & \textbf{4.072} & 10.060 & \textbf{6.856} & 13.376 & \textbf{4.196} & 7.799 \\
IS ($\uparrow$) & \textbf{9.194} & 8.870 & \textbf{10.348} & 9.874 & 2.424 & \textbf{2.535} \\
Precision ($\uparrow$) & 0.597 & \textbf{0.637} & 0.501 & \textbf{0.560} & 0.696 & \textbf{0.725} \\
Recall ($\uparrow$) & \textbf{0.558} & 0.416 & \textbf{0.485} & 0.452 & \textbf{0.539} & 0.524 \\
Density ($\uparrow$) & 0.999 & \textbf{1.702} & 0.925 & \textbf{1.196} & 1.368 & \textbf{1.552} \\
Coverage ($\uparrow$) & 0.962 & \textbf{0.992} & 0.947 & \textbf{0.972} & 0.984 & \textbf{0.990} \\
\hline
\end{tabular}
\label{table:2_ImprovedDDPM}
\end{table}

\begin{table}
\centering
\setlength{\tabcolsep}{2pt}
\renewcommand{\arraystretch}{1.3}
\caption{Evaluation of LDM vs $\text{I-Diff}_{\text{LDM}}$ for CIFAR10, CIFAR100, CelebA datasets}
\begin{tabular}{c|cc|cc|cc}
\hline
\multirow{2}{*}{\makecell[c]{Evaluation\\Metrics}} & \multicolumn{2}{c|}{CIFAR-10} & \multicolumn{2}{c|}{CIFAR-100} & \multicolumn{2}{c}{CelebA} \\
\cline{2-7}
& LDM & $\text{I-Diff}_{\text{LDM}}$ & LDM & $\text{I-Diff}_{\text{LDM}}$ & LDM & $\text{I-Diff}_{\text{LDM}}$ \\
\hline
FID ($\downarrow$) & \textbf{10.361} & 10.447 & \textbf{14.591} & 15.597 & \textbf{7.557} & 7.909 \\
IS ($\uparrow$) & 8.602 & \textbf{8.610} & 9.314 & \textbf{9.424} & \textbf{2.411} & 2.406 \\
Precision ($\uparrow$) & 0.674 & \textbf{0.712} & 0.662 & \textbf{0.680} & 0.773 & \textbf{0.776} \\
Recall ($\uparrow$) & \textbf{0.396} & 0.370 & \textbf{0.399} & 0.380 & 0.476 & 0.476 \\
Density ($\uparrow$) & 1.726 & \textbf{2.013} & 1.729 & \textbf{1.900} & 1.741 & \textbf{1.766} \\
Coverage ($\uparrow$) & 0.992 & \textbf{0.995} & 0.989 & \textbf{0.992} & 0.992 & \textbf{0.993} \\
\hline
\end{tabular}
\label{table:3_LDM}
\end{table}
\subsection{Performance Comparison between DDPM and I-Diff}
To compare the performance between DDPM and I-Diff, the modified loss function was utilized Oxford Flower dataset, Oxford IIIT Pet dataset, CIFAR-10 dataset, CIFAR-100 dataset, CelebA dataset, and CelebAHQ dataset. FID, IS along with Precision, Recall, Density, and Coverage were used to evaluate the quality of the generated samples of the image datasets.

Table \ref{table:1-DDPM_vs_IDiff} shows the comparison between DDPM and DDPM with the modified objective function based on trace mean (I-Diff) in terms of the generative model's evaluation metrics for the aforementioned image datasets. Across the datasets we observed that the fidelity metrics, Precision, Density as well Coverage have been improved in I-Diff compared to DDPM. Moreover, the results on Table \ref{table:2_ImprovedDDPM} and Table \ref{table:3_LDM} back up these improvements made by our modified loss by showing steady gains in fidelity metrics with another diffusion models such as Improved Diffusion \cite{nichol_improved_2021} as well as LDM \cite{rombach_high-resolution_2021} for image datasets. Furthermore, the results in Table VIII, IX, X, and XI in Supplementary Materials depict the improvement in fidelity metrics for DDPMs trained under different variance schedulers for 2D synthetic datasets. This proves that our method works well across multiple modes of operation where DDPM has been utilized. In addition, it confirmed the model-agnostic nature of our proposed method.
Although the FID and IS are considered to be the most widely used evaluation metrics for assessing image generation, we see that in the case of all datasets, they convey little to no discerning information about the generative ability of the proposed method and the original DDPM. But, by using other metrics such as the Precision, Recall, Density, and Coverage (PRDC), we can state that while our proposed method suffers a bit in terms of Recall, the generated samples,  are very close to being real (see Fig. \ref{fig:generated_images} in the paper, Fig. 6, 7, 8, 9, and 10 in the Supplementary Material), as indicated by the improvements in the Precision and Density metrics.

\subsection{Interpretation of the results of synthetic data}
\begin{figure*}[!t]
\centering
\includegraphics[width=1.0\textwidth]{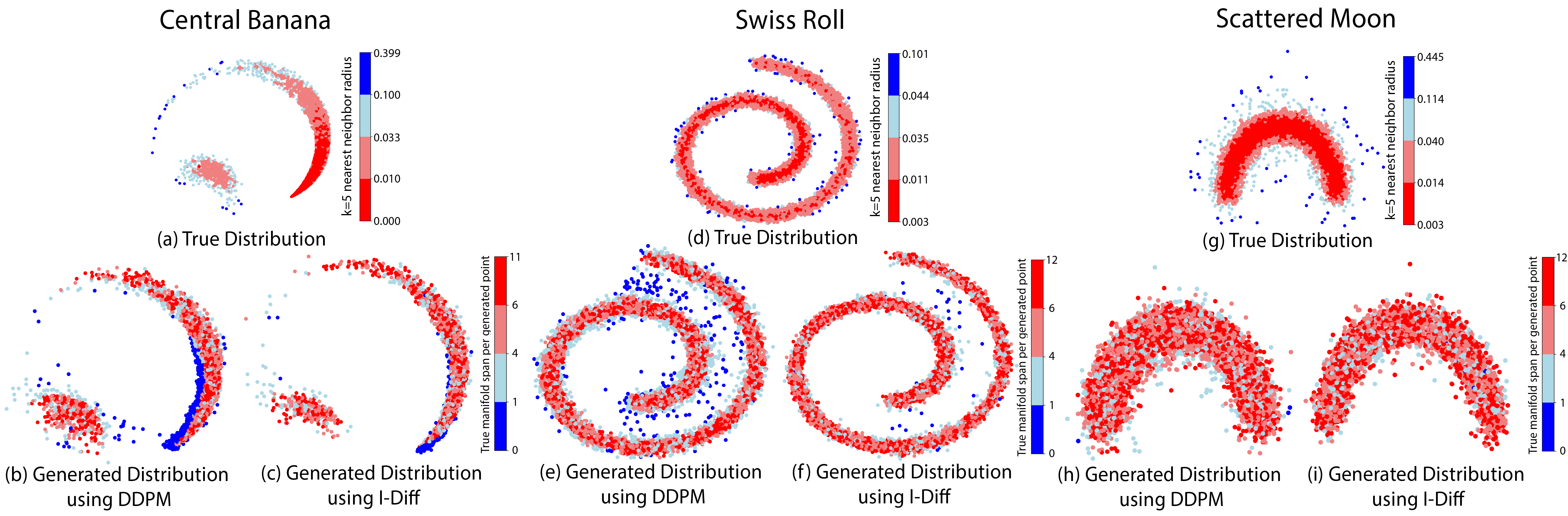}
\caption{\textbf{Central Banana, Swiss Roll, and Scattered Moon 2D synthetic datasets:} (a),(d),(g) True distribution points, color-coded by k=5 nearest neighbor radius. (b),(e),(h) DDPM-generated points, color-coded by true manifold span per point. (c),(f),(i) I-Diff generated points, color-coded by true manifold span per point.}
\label{fig:4}
\end{figure*}
We believe that the disparity in the changes of Precision, Recall, Density, and Coverage is a direct consequence of imposing a structural constraint on the objective function. It is evident that by focusing on the structure or the isotropy of the distribution, our method is capable of capturing highly dense mode regions and generating samples near them rather than being too diverse. Thus, it increases the fidelity but decreases the diversity of the generated samples.
As illustrated in the Fig. \ref{fig:4}(a), the Central Banana distribution was designed by introducing a distinct mode to the main structure of the distribution resulting in a multimodal distribution with a density gradient. Once, it is generated via I-Diff as indicated in Fig. \ref{fig:4}(c), it is evident that,
I-Diff, is capable of capturing the main structure even with the discontinuities of the density gradient. However, the illustrations show that DDPM lacks the capability of capturing the discontinuity in the density gradient between the tail end of the main distribution and the distinct mode. Instead, it tries to generate data points that are actually not even in the true distribution by interpolating along the main lobe's trend (see Fig. \ref{fig:4}(b)). Moreover, the limited capability to capture the discontinuity in the density gradient of DDPM can be further observed in the Swiss Roll distribution \ref{fig:4}(d) as well (see Fig. \ref{fig:4}(e) and \ref{fig:4}(f)). The increase in Density and decrease in Coverage for the datasets Swiss Roll and Central Banana are clear evidence for the aforementioned observations.  Hence, it is limited in ability to capture the underlying structure of the distribution. Additionally, there is a noticeable trend of generating data points (blue points in Fig. \ref{fig:4}(b), \ref{fig:4}(c)), outside the boundaries of the highly dense regions of the main lobe. This effect is likely due to the model's focus on these high-density regions. However, compared to DDPM, I-Diff effectively regulates the overgeneration of data points outside the boundaries of densely packed regions. This improvement is likely a result of the added regularization in the improved objective function, which encourages capturing the main structure of the true distribution. Moreover, this provides direct evidence that the introduced structural measure enables I-Diff to learn the density gradient of the target distribution more faithfully. As observed in Fig. \ref{fig:4}(c), fewer samples appear in the low-density tail of the main lobe compared to Fig. \ref{fig:4}(b), indicating improved control over sampling in sparse regions. This behavior reflects proper density-gradient matching rather than mere boundary alignment. Hence, it depicts the capability of I-Diff to create more representative distributions and thereby generate more representative samples of high precision and fidelity.
 
As illustrated in the Fig. \ref{fig:4}(g), the Scattered Moon distribution was designed by imposing scattered noise around the main structure of the data distribution. Once, it is generated via I-Diff as indicated in the Fig. \ref{fig:4}(i), it is evident that, the model has tried to only capture the prominent structure of the distribution without being susceptible to the low probable regions. Whereas, the DDPM model shows limitations in capturing the distinction between the main structure and the scattered noise (see Fig. \ref{fig:4}(h)). The increased Density and reduced Coverage values support this observation. This shows that the proposed objective function, enforces the generated samples to contain properties that push them to be closely linked to the real data. Thus, we can directly observe an improvement in the Density metric as it measures the sample fidelity. We believe that in the context of unconditional image generation, the isotropy based objective function helps the model learn to keep the generated samples closer to the high-density regions of the ground-truth distribution. Moreover, in the presence of corrupted training data, this tendency can also reduce the influence of such samples by favoring structurally consistent regions during generation.

These observations highlight the proposed algorithm's ability to increase Density by focusing on the dense regions of the true distribution. At the same time, the absence of generated data in the neighborhoods of low probable data points in the true distribution may result in a reduction in Coverage. When scattering is minimal, Coverage remains consistent. This indicates that the algorithm effectively captures the main structure of the true distribution without extending into low probable regions. This explains the percentage increase differences in different synthetic datasets. (See Fig. \ref{fig:objectiveFunction}) Particularly, in the Scattered Moon dataset, which consists of a higher amount of low-probability regions due to scattered data, the fidelity measure improvement is marginal compared to other datasets. Also, each of these metrics has its own utility depending on the application \cite{kynkaanniemi_improved_2019}. Thus, this should motivate the research community to propose new evaluation metrics such as Density, which is a much more meaningful measure of fidelity than FID and IS, to assess generative models.

Preserving the modality of a data distribution is essential, as failing to capture it can lead to a loss of structural details or edge information, both of which represent high-level features in computer vision and image processing tasks \cite{guo_density-aware_2020, shaham_singan_2019}

\subsection{Interpretation of the results of image data}
When the results in Table \ref{table:1-DDPM_vs_IDiff} are examined alongside the earlier discussion of PRDC metrics, it becomes clear that the proposed method increases the fidelity of the generated data, as indicated by higher Precision and Density, yet still yields comparatively high FID scores across all datasets. A similar trend is reported in the Improved DDPM \cite{nichol_improved_2021}, where introducing the regularizer $\text{L}_{\text{vlb}}$, which learns the variance of the reverse transition, leads to higher FID even though optimizing $\text{L}_{\text{vlb}}$ improves the log likelihood. In the same way, the proposed method raises fidelity but does so at the cost of lower Recall (a measure of diversity) and higher FID. This pattern is further reinforced by the substantial gains in Precision and Density observed for $\text{I-Diff}_{\text{Imp.}}$ (Improved DDPM with our regularizer) relative to Improved DDPM \cite{nichol_improved_2021}, though accompanied by a notable increase in FID.
Moreover, the metric Coverage indicates how much the generated data covers the true data. In each and every case, the models have been able to record high Coverage values across all datasets which confirms that the generated data has covered the vast majority of the true data. Therefore, it can be confirmed that the high Density values do not suggest a mode-seeking behavior in the true distribution, it actually learns the structure of the true distribution.

\section{Conclusion and Future Work}

In this work, we propose a simple, computationally inexpensive, carefully designed regularizer for the objective function of the conventional DDPM. The proposed model (I-Diff) has been able to improve the fidelity of the generated data across several benchmark datasets. Moreover, the improved fidelity of different variants of DDPMs suggests the framework-independent nature of the proposed method. This can be applied in a wide variety of disciplines where high fidelity and precise data generation are crucial, such as medical imaging, remote sensing, and speech synthesis. Moreover, this can be further applied to mitigate the effect of less precise training data, which would otherwise ultimately result in accumulated error in downstream tasks. This work also highlights the benefits of utilizing evaluation metrics beyond FID and IS, as Precision, Recall, Density, and Coverage provide additional insight into fidelity and diversity that traditional metrics may not fully reflect. We firmly believe that our work offers other researchers novel perspectives to explore even better regularizers to capture the structure more effectively. In future research, we aim to investigate domain-specific applications in the aforementioned areas, pairing them with our regularizer to further advance high-fidelity generation. 

\bibliographystyle{IEEEtran}
\bibliography{references}
\clearpage
\section{Supplementary Materials}

\subsection{Verifying the standard diffusion space and I-Diff space as Metric Spaces
\label{proof}}
\begin{equation}
    d_{L^2}(p, q) = \left[ \int (p(y) - q(y))^2 \, dy \right]^{\frac{1}{2}}
    \label{eq:s28}
\end{equation}

As mentioned in Section II-C, for the standard diffusion space, defined by the distance \( d_{L^2}(p, q) \) to qualify as a metric space, the distance must satisfy the following properties:

{\bf Non-negativity} \\
Since $p(y), q(y) \in \mathbb{R}$
\begin{align}
    (p(y) - q(y))^2 &\geq 0 \quad \text{; as } p(y), q(y) \in \mathbb{R} \nonumber \\
    \int (p(y) - q(y))^2 \, dy &\geq 0 \quad \nonumber\\
    \left[ \int (p(y) - q(y))^2 \, dy \right]^{\frac{1}{2}} &\geq 0 \nonumber
\end{align}

\begin{equation}
    \therefore d_{L^2}(p, q) \geq 0
    \label{eq:s29}
\end{equation}

\( d_{L^2}(p, q) = 0 \) if and only if \( p(y) = q(y) \) for every \( y \). Therefore, the \( d_{L^2} \) distance is zero only when the functions are essentially identical.

Hence, \( d_{L^2}(p, q)\) satisfies the non-negativity property.

{\bf Symmetry}
\begin{align}
d_{L^2}(p, q) &= \left[ \int (p(y) - q(y))^2 \, dy \right]^{1/2} \notag \\
&= \left[ \int (-(q(y) - p(y)))^2 \, dy \right]^{1/2} \notag \\
&= \left[ \int (q(y) - p(y))^2 \, dx \right]^{1/2} \nonumber \\
&= d_{L^2}(q, p) \nonumber \\
\therefore \quad d_{L^2}(p, q) &= d_{L^2}(q, p)
\label{eq:s31}
\end{align}

Hence, \( d_{L^2}(p, q)\) satisfies the symmetry property.

{\bf Triangle inequality}
\begin{align}
d_{L^2}(p, r) &= \left[ \int (p(y) - r(y))^2 \, dy \right]^{\frac{1}{2}} \notag \\
d_{L^2}^2(p, r) &= \int (p(y) - r(y))^2 \, dy \notag \\
&= \int \left( (p(y) - q(y)) + (q(y) - r(y)) \right)^2 dy \notag \\
&= \int (p(y) - q(y))^2 dy + \int (q(y) - r(y))^2 dy \notag \\
&\quad + 2 \int (p(y) - q(y))(q(y) - r(y)) \, dy
\label{eq:s32}
\end{align}

\noindent
However, by the Cauchy--Schwarz inequality:
\begin{align}
\left| \int (p(y) - q(y))(q(y) - r(y)) \, dy \right| \nonumber
\end{align}
\begin{align}
&\leq \sqrt{ \int (p(y) - q(y))^2 \, dy } \cdot \sqrt{ \int (q(y) - r(y))^2 \, dy } \notag \\
&= d_{L^2}(p, q) \cdot d_{L^2}(q, r)
\label{eq:s33}
\end{align}
By applying the result in (\ref{eq:s33}) to (\ref{eq:s32}), we obtain;
\begin{align}
2\int (p(y) - q(y))(q(y) - r(y)) \, dy 
&\leq 2 d_{L^2}(p, q) \cdot d_{L^2}(q, r) \nonumber
\label{eq:s34} \\
\therefore \quad d_{L^2}^2(p, r) 
&\leq d_{L^2}^2(p, q) + d_{L^2}^2(q, r) \nonumber \\
&+ 2 d_{L^2}(p, q) \cdot d_{L^2}(q, r) \nonumber \\
&= \left[ d_{L^2}(p, q) + d_{L^2}(q, r) \right]^2 \nonumber
\end{align}
Hence,
\begin{equation}
 \quad d_{L^2}(p, r) \leq d_{L^2}(p, q) + d_{L^2}(q, r)
\label{eq:s35}
\end{equation}
i.e. \( d_{L^2}(p, q)\) satifies the triangle inequality condition.
Thus, \( d_{L^2}(p, q)\) qualifies as a metric space.

Now, let us define the new distance measure with a structural regularizer as:
\begin{equation}
d_{\text{new}}(p, q) = d_{L^2}(p, q) + \lambda \cdot d_I(p, q)
\label{eq:s36}
\end{equation}
\noindent
Where:
\begin{align}
& d_{L^2}(p, q) = \left[ \int (p(y) - q(y))^2 \, dy \right]^{\frac{1}{2}} \text{and} \nonumber \\ 
& d_I(p, q) = |I(p) - I(q)| \nonumber
\end{align}
Here, \( I \) is the statistical measure mentioned in the Section II-C, and \( \lambda( > 0) \) is the regularization parameter.

As mentioned in Section II-C, for the new distance $d_{\text{new}}(p, q)$ in the I-Diff Space, to qualify as a metric space, the distance must satisfy the following properties:

{\bf Non-negativity} \\
According to the result in (\ref{eq:s29}), \( d_{L^2}(p, q) \geq 0 \). \\
Since \( I(p), I(q) \in \mathbb{R} \), we have \( |I(p) - I(q)| \geq 0 \). \(\therefore d_I(p, q) \geq 0\). \\
Thus, 
\[
d_{\text{new}}(p, q) \geq 0
\]
Hence, $d_{\text{new}}(p, q)$ satisfies the non-negativity property.

{\bf Symmetry} \\
According to (\ref{eq:s31}), \( d_{L^2}(p, q) = d_{L^2}(q, p) \). \\
Also, \( |I(p) - I(q)| = |I(q) - I(p)| \). \(\therefore d_I(p, q) = d_I(q, p)\). \\
Thus,
\[
d_{\text{new}}(p, q) = d_{\text{new}}(q, p)
\]
Hence, $d_{\text{new}}(p, q)$ satisfies the symmetry property.

{\bf Triangle inequality} \\
According to (\ref{eq:s35}), 

\begin{equation}
d_{L^2}(p, r) \leq d_{L^2}(p, q) + d_{L^2}(q, r)
\label{eq:s37}
\end{equation}

Since \( I(p), I(q), I(r) \in \mathbb{R} \), we have

\begin{align}
|I(p) - I(r)| 
&= |(I(p) - I(q)) + (I(q) - I(r))|  \nonumber \\
&\leq |I(p) - I(q)| + |I(q) - I(r)|
\label{eq:s38}
\end{align}

From (\ref{eq:s37}) and (\ref{eq:s38}), it follows that:

\begin{align}
d_{L^2}(p, r) + \lambda \cdot d_I(p, r) 
&\leq d_{L^2}(p, q) + \lambda \cdot d_I(p, q) \\
&+ d_{L^2}(q, r) + \lambda \cdot d_I(q, r) \\
\therefore \quad d_{\text{new}}(p, r) &\leq d_{\text{new}}(p, q) + d_{\text{new}}(q, r)
\end{align}
Hence, $d_{\text{new}}(p, q)$ satisfies the triangle inequality.

Thus, both the standard diffusion space with the \( d_{L^2}(p, q) \) distance and the new I-Diff space with the \( d_{\text{new}}(p, q) \) distance measure are metric spaces.
\vspace{-3mm}
\subsection{Training Configurations}
\label{sec:training_configurations}
This appendix provides a modular description of all training configurations used in our experiments. Unless otherwise stated, all models were trained using the Adam optimizer without weight decay. We used a diffusion process with 1000 timesteps and applied an exponential moving average (EMA) to model parameters, using decay factors in the range 0.9999. Sampling used the standard ancestral reverse-diffusion update. Gradient clipping was enabled for datasets with a maximum norm of 1.0. Dropout used was 0.1 for DDPM and 0.3 for Improved Diffusion. These global defaults form the baseline for the more detailed specifications below.

\noindent\textbf{Pixel-Space DDPMs}

All DDPMs follow the design outlined in the original DDPM (2020) [10] by Ho \textit{et al.} and share a common architectural structure unless a dataset-specific override is noted.

\subsubsection*{Architecture}

We used a ResNet-based U-Net with sinusoidal time embeddings. Attention layers were inserted at an intermediate spatial resolution (e.g., $16\times16$ for CIFAR and CelebA, Pet/Flower). Each resolution level contains two residual blocks. Channel multipliers vary by dataset:
\begin{itemize}
    \item CIFAR-10/100: $[1,2,2,2]$ with base width 128.
    \item Oxford-IIIT-Pet and Oxford Flower: $[1,2,4,8]$ with base width 64.
    \item CelebA ($64\times64$): $[1,2,2,2,4]$ with base width 128.
    \item CelebA-HQ ($256\times256$): $[1,1,2,2,4,4]$ with base width 128.
\end{itemize}

\subsubsection*{Diffusion Setup}

All DDPM models used the linear variance schedule with $\beta_1 = 10^{-4}$ and $\beta_T = 2\times 10^{-2}$. The U-Net predicts $\epsilon_\theta(x_t,t)$ and is trained using the standard noise-prediction surrogate objective.

\subsubsection*{Optimization}

Dataset-specific settings are as follows:
\begin{itemize}
    \item CIFAR10 and CIFAR100: batch size 128, learning rate $2\times10^{-4}$, 508K iterations.
    \item Oxford-IIIT-Pet: batch size 32, learning rate $2\times10^{-4}$, 256K iterations.
    \item Oxford Flower: batch size 32, learning rate $2\times10^{-4}$, 256K iterations.
    \item CelebA ($64\times64$): batch size 128, learning rate $2\times10^{-4}$, 500K iterations.
    \item CelebAHQ ($256\times256$): batch size 8, learning rate $2\times10^{-5}$, 5M iterations.
\end{itemize}
\noindent\textbf{Improved Diffusion Models}

Improved diffusion models follow the design of standard design used in Improved DDPM (2021) [18] by Nichol \textit{et al.}. They share the same U-Net backbones used for DDPM  but differ in two key aspects: a cosine noise schedule and a hybrid objective that learns both the predicted noise and the variance.

\subsubsection*{Architecture}

The core U-Net is identical to that used in DDPM, with the number of residual blocks increased when specified. Attention is inserted at the same spatial resolutions as in the DDPM counterparts.

\subsubsection*{Diffusion Setup}

All improved-diffusion models use the cosine variance schedule and the hybrid noise-prediction and learned-variance objective. The number of diffusion steps remains 1000.

\subsubsection*{Optimization}

\begin{itemize}
    \item CIFAR-10 and CIFAR-100: batch size 128, learning rate $1\times10^{-4}$, 300K iterations, 3 residual blocks per resolution.
    \item CelebA ($64\times64$): batch size 128, learning rate $1\times10^{-4}$, 500K iterations, 2 residual blocks per resolution.
\end{itemize}
\noindent\textbf{Latent Diffusion Models}

Latent diffusion models operate in a learned latent space rather than pixel space, following the standard design introduced in the Latent Diffusion Models framework [13] by Rombach \textit{et al.} (2021)

\subsubsection*{First-Stage Autoencoder}

LDM experiments use a trained AutoencoderKL to map $64\times64$ RGB images to a latent tensor of size $8\times8\times3$ for CelebA dataset and $64\times64$ RGB images to a latent tensor of size $32\times32\times3$ for CIFAR10 and CIFAR100. The encoder and decoder both use channel multipliers $[1,2,4,8]$ with two residual blocks per resolution. The latent prior is enforced via a KL divergence term. The autoencoder is trained separately and frozen for all diffusion experiments.

\subsubsection*{Latent U-Net}

The diffusion U-Net in latent space uses a base channel width of 128 for CIFAR10, CIFAR100 and 192 for CelebA with channel multipliers $[1,2,4]$. Attention is applied as latent resolutions $8$, $4$, and $2$. Two residual blocks are used at each resolution. Dropout is not applied in any LDM configuration.

\subsubsection*{Diffusion and Optimization}

The diffusion process uses a linear schedule between $\beta_{\text{start}} = 0.0015$ and $\beta_{\text{end}} = 0.0195$. The loss is the standard simple $L_2$ noise-prediction loss. Models were trained using the Adam optimizer with batch size 64 and a total of 500K iterations. A learning rate $2\times10^{-4}$ was used for all configuration.
\vspace{-5mm}
\subsection*{Algorithms}

Algorithm 1 and Algorithm 2 show the complete training and sampling procedure with the modified objective function for DDPM, including the isotropy regularizer.
\vspace{-3mm}
\begin{algorithm}[H]
\caption{Training Procedure}
\label{alg:training}
{\small
\begin{tabular}{p{0.92\columnwidth}}
 \textbf{repeat} \\
 \quad \( x_0 \sim q(x_0) \) \\
 \quad \( t \sim \mathrm{Uniform}(\{1,\ldots,T\}) \) \\
 \quad \( \epsilon \sim \mathcal{N}(0,\mathrm{I}) \) \\
 \quad Take gradient descent step on \\
\quad \(
\nabla_\theta \left( 
\mathbb{E}\|\epsilon - \epsilon_\theta(x_t,t)\|^2 
+ \lambda\left(\mathbb{E}[\epsilon_\theta^\top \epsilon_\theta / n] - 1\right)^2
\right)
\) \\
 \textbf{until} converged \\
\end{tabular}
}
\end{algorithm}
\vspace{-7mm}
\begin{algorithm}[H]
\caption{Sampling Procedure}
\label{alg:sampling}
{\small
\begin{tabular}{p{0.92\columnwidth}}
 \( x_T \sim \mathcal{N}(0,\mathrm{I}) \) \\
 \textbf{for} \( t = T,\ldots,1 \) \textbf{do} \\
 \quad \( z \sim \mathcal{N}(0,\mathrm{I}) \) if \(t>1\), else \(z=0\) \\
 \quad
\(
x_{t-1}
=
\frac{1}{\sqrt{\alpha_t}}
\left(
x_t
-
\frac{1-\alpha_t}{\sqrt{1-\bar{\alpha}_t}}\,\epsilon_\theta(x_t,t)
\right)
+ \sigma_t z
\) \\
 \textbf{end for} \\
 \textbf{return} \( x_0 \) \\
\end{tabular}
}
\end{algorithm}
\vspace{-9mm}
\subsection{Additional Results}
\vspace{-6pt}

\begin{table*}[!t]
    \centering
    \renewcommand{\arraystretch}{1.1}
    \caption{Precision variation across different objective functions for synthetic datasets}
    \begin{tabular}{c|c|c|c|c}
    \hline
         Objective Function & Central Banana & Moon Circles & Scattered Moon & Swiss Roll \\
         \hline
         Default & 0.974 & 0.991 & 0.999 & 0.983 \\
Mean & 0.978 (+0.411\%) & 0.994 (+0.303\%) & 0.999 (0.000\%) & 0.984 (+0.102\%) \\
Skewness & 0.974 (0.000\%) & 0.994 (+0.303\%) & 0.999 (0.000\%) & 0.985 (+0.203\%) \\
Kurtosis & 0.968 (-0.616\%) & 0.994 (+0.303\%) & 0.999 (0.000\%) & 0.975 (-0.814\%) \\
KL Divergence & 0.979 (+0.513\%) & \textbf{0.995 (+0.404\%)} & 0.999 (0.000\%) & 0.986 (+0.305\%) \\
Maximum Mean Discrepancy & 0.980 (+0.616\%) & 0.994 (+0.303\%) & 0.999 (0.000\%) & 0.986 (+0.305\%) \\
Iso-Trace Mean & \textbf{0.984 (+1.027\%)} & \textbf{0.995 (+0.404\%)} & \textbf{1.000 (+0.100\%)} & \textbf{0.987 (+0.407\%)} \\
Iso-Frobenius Norm & 0.977 (+0.308\%) & 0.993 (+0.202\%) & 0.999 (0.000\%) & 0.977 (-0.610\%) \\
Iso-Diagonal Split & 0.983 (+0.924\%) & \textbf{0.995 (+0.404\%)} & \textbf{1.000 (+0.100\%)} & 0.979 (-0.407\%) \\
Iso-Log Eigenvalue Penalty & 0.981 (+0.719\%) & 0.994 (+0.303\%) & 0.999 (0.000\%) & 0.983 (0.000\%) \\
Iso-Bures Distance Penalty & 0.977 (+0.308\%) & 0.994 (+0.303\%) & 0.999 (0.000\%) & 0.979 (-0.407\%) \\
\hline
    \end{tabular}
    \label{tab:ObjectiveFunction_Precision}
\end{table*}
\vspace*{-10mm}
\begin{table*}[!t]
    \centering
    \renewcommand{\arraystretch}{1.1}
    \caption{Recall variation across different objective functions for synthetic datasets}
    \begin{tabular}{c|c|c|c|c}
    \hline
         Objective Function & Central Banana & Moon Circles & Scattered Moon & Swiss Roll \\
         \hline
         Default & \textbf{0.998} & \textbf{0.998} & 0.996 & \textbf{0.998} \\
Mean & 0.998 (0.000\%) & 0.998 (0.000\%) & 0.996 (0.000\%) & 0.996 (-0.200\%) \\
Skewness & 0.998 (0.000\%) & 0.997 (-0.100\%) & 0.996 (0.000\%) & 0.997 (-0.100\%) \\
Kurtosis & 0.998 (0.000\%) & 0.998 (0.000\%) & \textbf{0.998 (+0.201\%)} & 0.998 (0.000\%) \\
KL Divergence & 0.998 (0.000\%) & 0.997 (-0.100\%) & 0.996 (0.000\%) & 0.995 (-0.301\%) \\
Maximum Mean Discrepancy & 0.998 (0.000\%) & 0.997 (-0.100\%) & 0.996 (0.000\%) & 0.996 (-0.200\%) \\
Iso-Trace Mean & 0.997 (-0.100\%) & 0.997 (-0.100\%) & 0.996 (0.000\%) & 0.994 (-0.401\%) \\
Iso-Frobenius Norm & 0.997 (-0.100\%) & 0.998 (0.000\%) & 0.995 (-0.100\%) & 0.998 (0.000\%) \\
Iso-Diagonal Split & 0.998 (0.000\%) & 0.998 (0.000\%) & 0.993 (-0.301\%) & 0.997 (-0.100\%) \\
Iso-Log Eigenvalue Penalty & 0.998 (0.000\%) & 0.998 (0.000\%) & 0.994 (-0.201\%) & 0.994 (-0.401\%) \\
Iso-Bures Distance Penalty & 0.998 (0.000\%) & 0.998 (0.000\%) & 0.995 (-0.100\%) & 0.997 (-0.100\%) \\
    \hline
    \end{tabular}
    \label{tab:ObjectiveFunction_Recall}
\end{table*}
\vspace*{-10mm}
\begin{table*}[!t]
    \centering
    \renewcommand{\arraystretch}{1.1}
    \caption{Density variation across different objective functions for synthetic datasets}
    \begin{tabular}{c|c|c|c|c}
    \hline
         Objective Function & Central Banana & Moon Circles & Scattered Moon & Swiss Roll \\
         \hline
         Default & 0.967 & 0.993 & 1.006 & 0.978 \\
Mean & 0.972 (+0.517\%) & 0.997 (+0.403\%) & 1.001 (-0.497\%) & 0.978 (0.000\%) \\
Skewness & 0.970 (+0.310\%) & 1.001 (+0.806\%) & 1.002 (-0.398\%) & 0.976 (-0.204\%) \\
Kurtosis & 0.958 (-0.931\%) & 0.998 (+0.504\%) & 1.003 (-0.298\%) & 0.956 (-2.249\%) \\
KL Divergence & 0.976 (+0.931\%) & \textbf{1.009 (+1.611\%)} & 1.001 (-0.497\%) & \textbf{0.989 (+1.125\%)} \\
Maximum Mean Discrepancy & 0.973 (+0.620\%) & 0.999 (+0.604\%) & 1.000 (-0.596\%) & 0.978 (0.000\%) \\
Iso-Trace Mean & 0.976 (+0.931\%) & 1.005 (+1.208\%) & \textbf{1.007 (+0.099\%)} & 0.987 (+0.920\%) \\
Iso-Frobenius Norm & 0.973 (+0.620\%) & 0.993 (0.000\%) & 1.004 (-0.199\%) & 0.963 (-1.534\%) \\
Iso-Diagonal Split & \textbf{0.984 (+1.758\%)} & 0.991 (-0.201\%) & 1.003 (-0.298\%) & 0.975 (-0.307\%) \\
Iso-Log Eigenvalue Penalty & 0.981 (+1.448\%) & 0.995 (+0.201\%) & 1.002 (-0.398\%) & 0.986 (+0.818\%) \\
Iso-Bures Distance Penalty & 0.977 (+1.034\%) & 0.994 (+0.101\%) & 1.000 (-0.596\%) & 0.972 (-0.613\%) \\
    \hline
    \end{tabular}
    \label{tab:ObjectiveFunction_Density}
\end{table*}
\vspace*{-10mm}
\begin{table*}[!t]
    \centering
    \renewcommand{\arraystretch}{1.1}
    \caption{Coverage variation across different objective functions for synthetic datasets}
    
    \begin{tabular}{c|c|c|c|c}
    \hline
    Objective Function & Central Banana & Moon Circles & Scattered Moon & Swiss Roll \\
    \hline
    Default & \textbf{0.963} & \textbf{0.964} & 0.964 & 0.953 \\
Mean & 0.956 (-0.727\%) & 0.959 (-0.519\%) & 0.966 (+0.207\%) & 0.941 (-1.259\%) \\
Skewness & 0.958 (-0.519\%) & 0.956 (-0.830\%) & 0.962 (-0.207\%) & \textbf{0.956 (+0.315\%)} \\
Kurtosis & 0.956 (-0.727\%) & 0.957 (-0.726\%) & 0.965 (+0.104\%) & 0.944 (-0.944\%) \\
KL Divergence & 0.955 (-0.831\%) & 0.961 (-0.311\%) & 0.966 (+0.207\%) & 0.950 (-0.315\%) \\
Maximum Mean Discrepancy & 0.956 (-0.727\%) & 0.961 (-0.311\%) & 0.965 (+0.104\%) & 0.953 (0.000\%) \\
Iso-Trace Mean & 0.953 (-1.038\%) & 0.961 (-0.311\%) & 0.966 (+0.207\%) & 0.947 (-0.630\%) \\
Iso-Frobenius Norm & 0.950 (-1.350\%) & 0.960 (-0.415\%) & \textbf{0.970 (+0.622\%)} & 0.935 (-1.889\%) \\
Iso-Diagonal Split & 0.955 (-0.831\%) & 0.957 (-0.726\%) & 0.969 (+0.519\%) & 0.928 (-2.623\%) \\
Iso-Log Eigenvalue Penalty & 0.955 (-0.831\%) & 0.953 (-1.141\%) & 0.966 (+0.207\%) & 0.928 (-2.623\%) \\
Iso-Bures Distance Penalty & 0.946 (-1.765\%) & 0.959 (-0.519\%) & 0.968 (+0.415\%) & 0.922 (-3.253\%) \\
    \hline
    \end{tabular}
    \label{tab:ObjectiveFunction_Coverage}
\end{table*}

\begin{table*}[!t]
    \centering
    \renewcommand{\arraystretch}{1.1}
    \caption{Variation of evaluation metrics with variance schedules for Central~Banana dataset}
    \begin{tabular}{c|cc|cc|cc|cc}
    \hline
    \multirow{2}{*}{Metrics} & \multicolumn{2}{c|}{Cosine Schedule} & \multicolumn{2}{c|}{Linear Schedule} & \multicolumn{2}{c|}{Quadratic Schedule} & \multicolumn{2}{c}{Sigmoid Schedule} \\
\cline{2-9}
& DDPM & I-Diff & DDPM & I-Diff & DDPM & I-Diff & DDPM & I-Diff \\
\hline
    Precision & 0.971 & \textbf{0.990} & 0.970 & \textbf{0.984} & 0.976 & \textbf{0.988} & 0.976 & \textbf{0.990}\\
    Recall & \textbf{0.998} & 0.997 & 0.998 & \textbf{0.999} & \textbf{0.999} & 0.997 & \textbf{0.999} & 0.996\\
    Density & 0.960 & \textbf{0.987} & 0.964 & \textbf{0.977} & 0.974 & \textbf{0.991} & 0.966 & \textbf{0.993}\\
    Coverage & \textbf{0.953} & 0.952 & \textbf{0.961} & 0.954 & \textbf{0.959} & 0.954 & 0.949 & 0.949\\
    \hline

    \end{tabular}
    \label{table:variance_central_banana}
\end{table*}

\begin{table*}[!t]
    \centering
    \renewcommand{\arraystretch}{1.1}
    \caption{Variation of evaluation metrics with variance schedules for Moon~Circles dataset}
    \begin{tabular}{c|cc|cc|cc|cc}
    \hline
    \multirow{2}{*}{Metrics} & \multicolumn{2}{c|}{Cosine Schedule} & \multicolumn{2}{c|}{Linear Schedule} & \multicolumn{2}{c|}{Quadratic Schedule} & \multicolumn{2}{c}{Sigmoid Schedule} \\
\cline{2-9}
& DDPM & I-Diff & DDPM & I-Diff & DDPM & I-Diff & DDPM & I-Diff \\
\hline
    Precision & 0.996 & \textbf{0.998} & 0.993 & \textbf{0.995} & 0.997 & \textbf{0.998} & 0.997 & \textbf{0.998}\\
    Recall & \textbf{0.998} & 0.995 & 0.998 & \textbf{0.999} & \textbf{0.997} & 0.995 & 0.997 & 0.997\\
    Density & 1.010 & \textbf{1.012} & 0.997 & \textbf{1.007} & 1.006 & \textbf{1.011} & 1.009 & \textbf{1.015}\\
    Coverage & \textbf{0.966} & 0.960 & \textbf{0.964} & 0.963 & 0.965 & 0.965 & 0.961 & \textbf{0.963}\\
    \hline
    \end{tabular}
    \label{table:variance_moonCircles}
\end{table*}

\begin{table*}[!t]
    \centering
    \renewcommand{\arraystretch}{1.1}
    \caption{Variation of evaluation metrics with variance schedules for Scattered~Moon dataset}
    \begin{tabular}{c|cc|cc|cc|cc}
    \hline
    \multirow{2}{*}{Metrics} & \multicolumn{2}{c|}{Cosine Schedule} & \multicolumn{2}{c|}{Linear Schedule} & \multicolumn{2}{c|}{Quadratic Schedule} & \multicolumn{2}{c}{Sigmoid Schedule} \\
\cline{2-9}
& DDPM & I-Diff & DDPM & I-Diff & DDPM & I-Diff & DDPM & I-Diff \\
\hline
    Precision & 0.999 & 0.999 & 0.999 & 0.999 & 0.999 & \textbf{1.000} & 0.999 & \textbf{1.000}\\
    Recall & \textbf{0.997} & 0.996 & \textbf{0.997} & 0.995 & \textbf{0.998} & 0.997 & \textbf{0.997} & 0.996\\
    Density & 1.001 & \textbf{1.010} & 1.004 & \textbf{1.006} & 1.001 & \textbf{1.006} & 0.998 & \textbf{1.009}\\
    Coverage & 0.967 & 0.967 & \textbf{0.966} & 0.963 & \textbf{0.967} & 0.966 & \textbf{0.965} & 0.963\\
    \hline
    \end{tabular}
    \label{table:variance_scatteredMoon}
\end{table*}

\begin{table*}[!t]
    \centering
    \renewcommand{\arraystretch}{1.1}
    \caption{Variation of evaluation metrics with variance schedules for Swiss~Roll dataset}
    \begin{tabular}{c|cc|cc|cc|cc}
    \hline
    \multirow{2}{*}{Metrics} & \multicolumn{2}{c|}{Cosine Schedule} & \multicolumn{2}{c|}{Linear Schedule} & \multicolumn{2}{c|}{Quadratic Schedule} & \multicolumn{2}{c}{Sigmoid Schedule} \\
\cline{2-9}
& DDPM & I-Diff & DDPM & I-Diff & DDPM & I-Diff & DDPM & I-Diff \\
\hline
    Precision & 0.990 & \textbf{0.994} & 0.983 & \textbf{0.986} & 0.990 & \textbf{0.993} & 0.991 & \textbf{0.993}\\
    Recall & \textbf{0.998} & 0.995 & \textbf{0.999} & 0.997 & \textbf{0.997} & 0.996 & \textbf{0.999} & 0.991\\
    Density & 0.991 & \textbf{1.009} & 0.965 & \textbf{0.989} & 0.984 & \textbf{1.012} & 0.990 & \textbf{1.010}\\
    Coverage & \textbf{0.961} & 0.960 & 0.944 & \textbf{0.953} & \textbf{0.959} & 0.951 & \textbf{0.966} & 0.954\\
    \hline
    \end{tabular}
    \label{table:variance_swissRoll}
\end{table*}

\begin{figure*}[!t]
\centering
\includegraphics[width=1.0\linewidth]{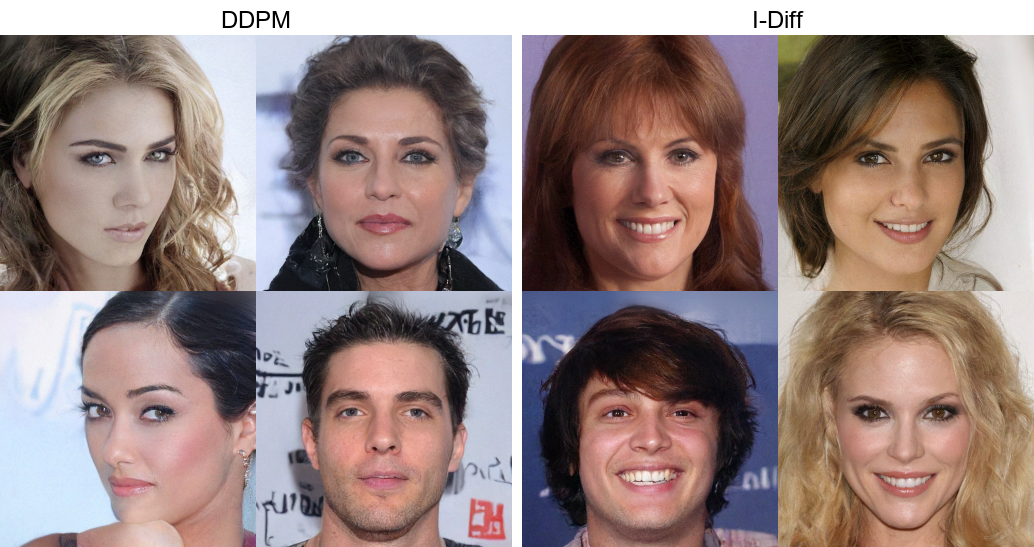}
\caption{\textbf{Generated images of the DDPM (left) and I-Diff (right)} for the CelebAHQ dataset.}
\label{fig:generated_images_celebahq}
\end{figure*}

\begin{figure*}[!t]
\centering
\includegraphics[width=1.0\linewidth]{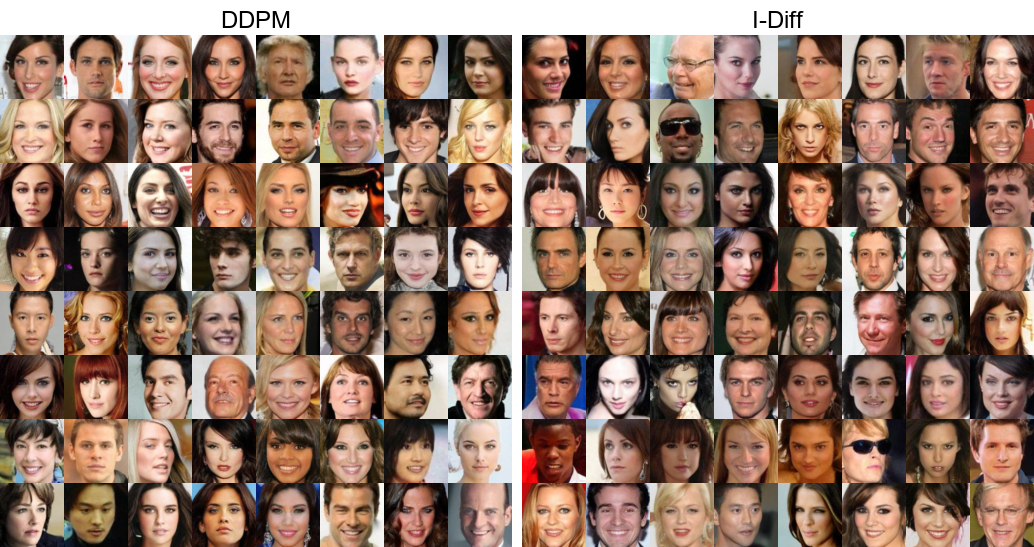}
\caption{\textbf{Generated images of the DDPM (left) and I-Diff (right)} for the CelebA dataset.}
\label{fig:generated_images_celeba}
\end{figure*}

\begin{figure*}[!t]
\centering
\includegraphics[width=1.0\linewidth]{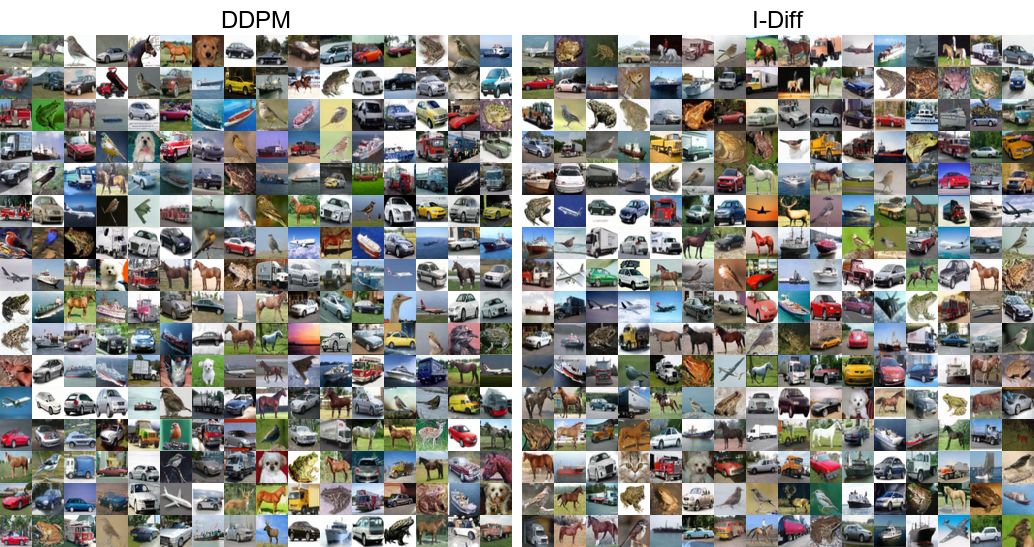}
\caption{\textbf{Comparison of the generated images via the DDPM (left) and I-Diff (right)} for the CIFAR10 dataset.}
\label{fig:generated_images_cifar10}
\end{figure*}

\begin{figure*}[!t]
\centering
\includegraphics[width=1.0\linewidth]{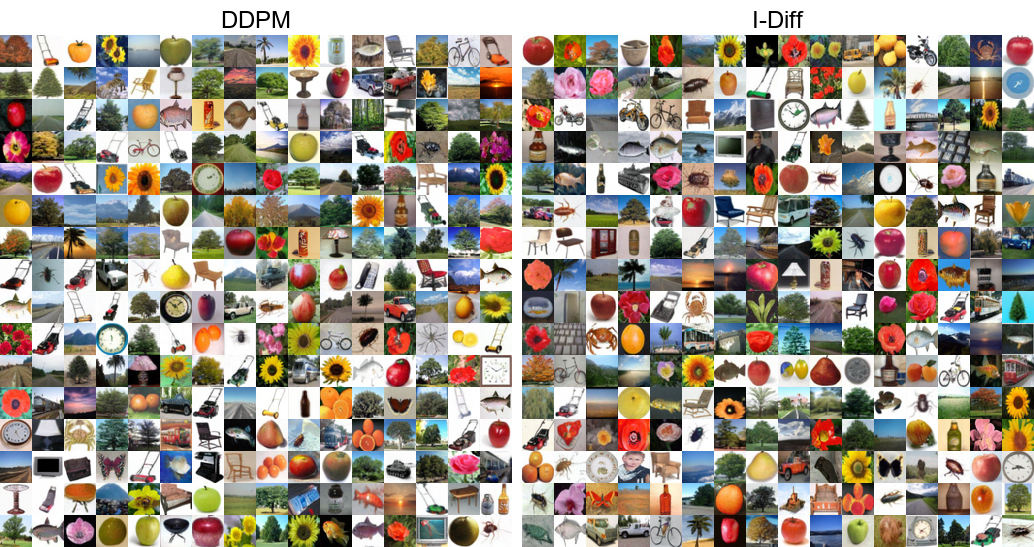}
\caption{\textbf{Generated images of the DDPM (left) and I-Diff (right)} for the CIFAR100 dataset.}
\label{fig:generated_images_cifar100}
\end{figure*}

\begin{figure*}[!t]
\centering
\includegraphics[width=1.0\linewidth]{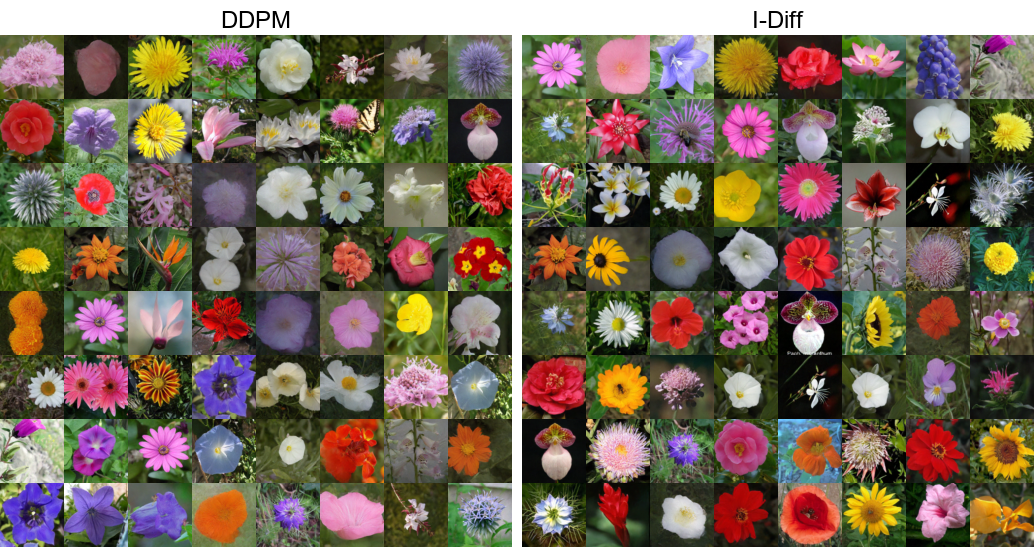}
\caption{\textbf{Generated images of the DDPM (left) and I-Diff (right)} for the Oxford Flowers dataset.}
\label{fig:generated_images_oxfordFlowers}
\end{figure*}

\end{document}